# Translating Dietary Standards into Healthy Meals with Minimal Substitutions

Trevor Chan[1,2,3] and Ilias Tagkopoulos[1,2,3]*

[1]Department of Computer Science, University of California, Davis, Davis, California 95616, USA

[2]Genome Center, University of California, Davis, Davis, California, 95616, USA

[3]USDA/NSF AI Institute for Next Generation Food Systems (AIFS), University of California, Davis, Davis, California, 95616 USA

*Corresponding author (itagkopoulos@ucdavis.edu)

**Keywords:** diet recommendation, public health, machine learning, nutrition, meal generation

## ABSTRACT

An important goal for personalized diet systems is to improve nutritional quality without compromising convenience or affordability. We present an end-to-end framework that converts dietary standards into complete meals with minimal change. Using the What We Eat in America (WWEIA) intake data for 135,491 meals, we identify 34 interpretable meal archetypes that we then use to condition a generative model and a portion predictor to meet USDA nutritional targets. In comparisons within archetypes, generated meals are better at following recommended daily intake (RDI) targets by 47.0%, while remaining compositionally close to real meals. Our results show that by allowing one to three food substitutions, we were able to create meals that were 10% more nutritious, while reducing costs 19-32%, on average. By turning dietary guidelines into realistic, budget-aware meals and simple swaps, this framework can underpin clinical decision support, public-health programs, and consumer apps that deliver scalable, equitable improvements in everyday nutrition.

## INTRODUCTION

Diet is one of the most powerful, modifiable drivers of obesity, diabetes, cardiovascular disease, and other non-communicable conditions, yet translating nutrition science into day-to-day meals remains difficult for most people[1]. Personalized diet recommendation systems promise scale and individualization, but many tools still optimize a single goal (taste, calories, or convenience), lack strict standards-based evaluation, and provide limited guidance on how to change as little as possible to eat better[2]. Consequently, there remains a gap between guideline-concordant diets and what recommenders reliably generate in practice[2]. Rule-based and expert-curated systems helped encode guidelines but often sacrificed adaptability and user fit[3]. Subsequent machine-learning approaches - clinical optimization for chronic disease (e.g., DietOS) and IoMT-assisted personalization - improved targeting, yet frequently treated health metrics in isolation and rarely reported controlled benchmarking against USDA nutrient standards[3,4]. Many-objective/evolutionary methods began balancing adequacy, preferences, and diversity, and clustering/classification pipelines introduced segmentation, but most do not close the loop from what to eat (composition) to how much to eat (portions), a tradeoff which is a key determinant of adequacy, balance, and moderation in real meals[5,6] and knowledge graph, health-aware recommenders[7]. Generative modeling has accelerated progress towards this goal. Systems, such as Yum-Me, explicitly model both nutrient goals and taste[8], while recent pipelines leveraging methods like variational autoencoders produce plausible meal plans[9]. More prevalent LLM-based approaches also explore more interactive suggestions[10]. However, systematic reviews highlight inconsistency and factual errors in LLM-generated nutrition, underscoring the need for domain constraints and transparent, multi-objective evaluation[11–14]. For real-world impact in nutrition science and clinical informatics, tools must embed dietary standards, quantify uncertainty, and deliver actionable and minimal-change recommendations that users can implement without overhauling habits[15,16].

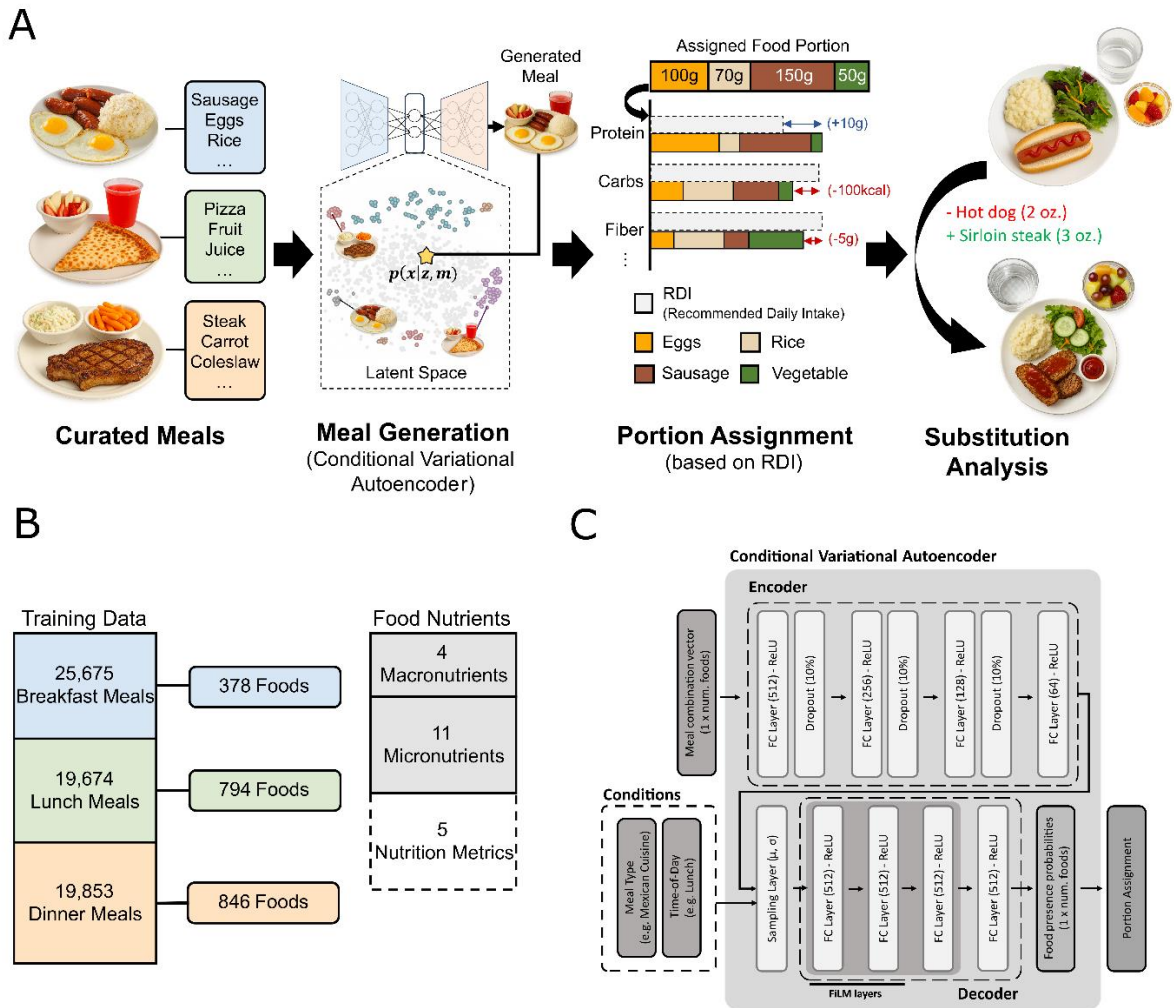

**Figure 1 | End-to-end meal generation, RDI-aware portioning, and substitution evaluation.**
(A) We start from curated meals and train a conditional variational autoencoder (CVAE) that samples realistic food combinations for a chosen archetype (e.g., breakfast) from a structured latent space. A portion assigner initializes standard servings and then adjusts grams to meet USDA RDI/AMDR targets while preserving the combination. Downstream, we evaluate minimal-change substitutions by searching a learned replaceability graph to find swaps that improve nutrition at lower or comparable cost under a portion-based restaurant pricing model. (B) Data and nutrients used for training and evaluation. Counts of meals per time-of-day, the number of foods available for each meal type, and the nutrient panel (4 macronutrients, 11 micronutrients, and 5 diet quality metrics). (C) The encoder/decoder of the CVAE are conditioned on meal type and calorie band via Feature-wise Linear Modulation (FiLM) layers. The decoder outputs food-presence probabilities which feed the portion assignment module.

Here, we present an end-to-end framework for meal generation, and substitution (**Figure 1A**) designed for public-health impact. We use a training set of 65,202 meals, consisting of 2,019 foods (1,475 of them unique), each associated with 20 nutrients and other nutrient-metrics (**Figure 1B**), and we partition them into 34 archetypes (e.g. protein and grains) based on food categories and nutritional content. Then we train a Conditional Variational Autoencoder (CVAE), conditioned to these archetypes to generate representative meals that are subsequently compared to real meals (**Figure 1C**).

## METHODS

**Data description.** The USDA "What We Eat in America" (WWEIA) component of the National Health and Nutrition Examination Survey (NHANES) was the primary source for all analyses[17–20]. We used six survey waves (2013–2020), comprising 55,228 respondents and 135,491 meals (**Supplementary Table 1**). WWEIA provides a hierarchical taxonomy of ingredients, foods, and meals indexed by USDA codes. Foods are constructed from ingredients and meals are constructed from foods. We standardized food codes across survey waves using USDA's discontinuation and renumbering documentation and retained dropped or revised codes where mapping was unambiguous. The final corpus contained 8,650 food codes and 2,940 ingredient codes. We excluded pre-2013 surveys because discontinuation mappings were incomplete, preventing reliable code harmonization. Definitions for terms used throughout are given in **Supplementary Table 2**. The pseudocode for the complete framework, which includes the data preprocessing, meal clustering, conditional meal generation, RDI portion assignment, and substitution optimization is provided in the Section 4 of the **Supplementary Material**.

**Data Processing.** The WWEIA dataset underwent comprehensive preprocessing to ensure data quality and reduce dimensionality. For surveys with discontinued food code documentation, we created a comprehensive dataset based on the discontinuation code given (**Supplementary Table 3**). For expanded, consolidated, and renumbered food codes, all previous food codes affected were replaced with the most recent food codes. Dropped and revised food codes were kept in the dataset. This yielded a dataset with 120,375 meals and 6,212 foods. Of these meals, 39,749 were breakfast meals, 37,397 were lunch meals, and 43,229 were dinner meals. Data was organized into time-of-day meal-to-food data subsets with corresponding gram amounts. These datasets were further reduced to binary data that represented the presence or absence of food per time-of-day meal. Local Outlier Factorization (LOF)[21] was conducted and the top-0.3 percent of foods were removed based on negative outlier factorization score, resulting in the removal of 120 breakfast meals, 113 lunch meals and 130 dinner meals. To address the sparsity presented in the dataset, i.e. many foods but only several foods per meal, we created a prototype mapping system using a nutrient-aware aggregation algorithm (**Supplementary Section 1.1**). This approach reduced the food space by 87.5% while maintaining nutritional representation, enabling more effective model training. Finally, we constructed bootstrap confidence intervals for the mean of each food in each meal-to-food data subset to reduce the list of ingredients to only those ingredients that were representative of the set of foods consumed. The lower bound of the confidence interval was used as a threshold for ingredient removal. This resulted in the filtered data subsets of 528, 627, and 686 foods to 39,435, 37,024, and 42,831 meals for breakfast, lunch, and dinner, respectively.

**Clustering Analysis.** We clustered meals separately by time-of-day (breakfast, lunch, dinner) to preserve meal-specific structure. Each meal was embedded in a hybrid, high-dimensional feature space that combined: (i) nutritional composition (macronutrients, fiber, energy, macro balance/density ratios) and (ii) categorical food composition (gram amounts per WWEIA main and subcategories, e.g., grains, fruits, vegetables, dairy, mixed dishes, snacks/sweets, beverages). Features were z-score standardized within meal type. We then applied an enhanced HDBSCAN[22,23] procedure with meal-type-specific parameters and post-processing cluster merging for stability and coverage. To profile clusters, we contrasted within-cluster vs. complement feature means and controlled multiple testing with Benjamini–Hochberg correction[24]. We required an absolute mean difference $\geq 0.10$ for significance and labeled features "distinctive" at $|\Delta| \geq 0.20$, reporting Cohen's d for effect sizes[25]. Detailed feature extraction, clustering parameters, statistical validation procedures, and cluster naming algorithms are described in **Supplementary Section 1.2**. This process produced stable, interpretable cluster archetypes for each meal type that were used as conditioning variables for generation and evaluation.

**Meal Generation.** Meal generation consisted of two stages: (1) a CVAE[26] that generates food combinations conditioned on meal clusters and meal types (breakfast/lunch/dinner), and (2) a simple RDI-per-kcal portion assignment that converts presence probabilities to portions for USDA-aligned reporting while preserving the model's food combinations. In the case of CVAE, the model uses dual conditioning on cluster and meal type (see **Supplementary Information** for architecture and hyperparameter choices).

**Portion Assignment.** Meals are generated directly in gram units using an RDI-per-kilocalorie strategy implemented within the generator module. Each meal type (breakfast, lunch, dinner) is assigned 25%, 35%, and 40% of a 2,000-kcal daily target, respectively (see **Supplementary Table 8**). The generator computes per-kcal nutrient densities from the daily RDI vector and optimizes food portions to minimize signed $\log_2$ deviations from these per-kcal targets while enforcing total-energy balance and realistic gram-level constraints ($\leq 900$ g per meal, $\leq 25\%$ kcal from beverages). Additional soft caps limit unrealistic component masses (**Supplementary Table 9**).

**Evaluation framework.** We conducted a head-to-head evaluation of generated meals across 19,013 meals (6,268 breakfast, 6,393 lunch, and 6,352 dinner meals per significant clusters in **Supplementary Table 6**), assessing established nutritional adequacy and diversity metrics[27–32]. **Table 1** shows the metrics used in greater detail. For analysis, Cohen's d was used for effect sizes. Bootstrap resampling (200 iterations) generated confidence intervals. FDR correction was applied for multiple comparisons. Model stability was assessed by 5-fold cross-validation.

| Metric | Definition | Formula | Notes |
| --- | --- | --- | --- |
| Mean Excess Ratio (MER) | Quantifies average proportional excess of overconsumed nutrients (e.g., sodium, saturated fat, added sugars) relative to upper dietary limits | $\frac{1}{K}\sum_{k=1}^{K} \frac{I_k}{L_k}$ | $I_k$: intake of nutrient $k$; $L_k$: recommended upper limit |
| Mean Adequacy Ratio (MAR) | Represents the average adequacy across essential micronutrients $n$, assessing overall nutrient sufficiency | $\frac{1}{N}\sum_{n=1}^{N} \text{NAR}_n$ | $N = 11$ micronutrients (Ca, Fe, Zn, Vitamins A, C, B6, B12, Thiamin, Riboflavin, Niacin, Folate) $\text{NAR}_n = \min(1, \frac{I_n}{\text{RDI}_n})$ |
| AMDR Composite | Evaluates macronutrient (protein, fat, carbohydrates) energy balance within acceptable macronutrient distribution ranges (AMDR) | $\frac{1}{M}\sum_{m=1}^{M} 1(L_m \leq E_m \leq U_m)$ | $E_m$: percent of total energy from macronutrient $m$; $L_m, U_m$: AMDR bounds (Protein 10–35%, Fat 20–35%, Carbohydrate 45–65%) |
| Diversity | Measures dietary variety across food groups $F$ using the Hill index; higher values indicate broader food-group representation | $\left(\sum_{i=1}^{F} p_i^q\right)^{\frac{1}{1-q}}$ | $p_i$: proportion of food group $i$; for $q=1$, $\exp(-\sum_i p_i \ln p_i)$ (Shannon diversity) |
| Energy Density | Represents caloric content per gram of meal, reflecting portion appropriateness and caloric concentration. | $\frac{\text{Total Calories}}{\text{Total Mass (g)}}$ | Lower values correspond to less calorically dense and typically more balanced meals. |

**Table 1 | Primary evaluation metrics for generated meal assessment**. Summary definitions and mathematical forms for the five principal metrics used to evaluate generated meals: Mean Excess Ratio (MER), Mean Adequacy Ratio (MAR), AMDR composite, dietary Diversity (Hill index), and Energy Density.

**Meal substitution analysis.** For each generated meal, we identify a small set of similar real meals to serve as candidate substitutions. Similarity is based on overall item overlap and composition, while requiring comparable meal energy and a similar number of items. We also consider simple single-item replacements within the same food category. We define k as the number of items replaced. For each candidate, we compute nutritional improvement as the reduction in average absolute deviation from per-meal Recommended Daily Intake (RDI) targets (25%/35%/40% of daily RDI for breakfast/lunch/dinner). Cost change is calculated from a price-per-100g map derived from retail price listings and category-specific multipliers are applied. We select winning substitutions using a simple trade-off between improvements in nutritional adequacy and cost increases, with optional budget and "no cost increase" constraints. Portions in candidate real-meal matches are taken as observed; for single-item swaps, the grams of the removed item are reassigned to the added item.

## RESULTS

**Meal clusters capture core U.S. meal archetypes across breakfast, lunch, and dinner.**

We retained 34 interpretable clusters that map onto common U.S. meal archetypes (i.e. general meal patterns) and differ meaningfully in nutrition (**Figure 2**). Several clusters reveal behavioral trade-offs. For example a lunch breads & spreads pattern (**Figure 2B**) pairs breads with a very high fruit share (fruit ratio *Cohen's d* = +12.41, $q < 10^{-271}$) and greater macro-diversity (*Cohen's d* = +1.97, $q < 10^{-114}$) but scores lower on overall meal balance (d = −2.05, $q < 10^{-103}$), suggesting that meals within have more snack-like plates rather than balanced entrée plates. Conversely, Mexican entrées concentrate on the main dish and trim the sides (ingredient count *Cohen's d* = −0.74, $q < 10^{-52}$; portion variability d = −0.66, $q < 10^{-37}$), with strong mixed-dish enrichment (*Cohen's d* = +4.47, $q < 10^{-300}$). Together, these nutrition-anchored contrasts show that the retained clusters for training represent robust meal archetypes spanning energy-dense plates (pizza dinners, sandwich lunches, cereal breakfasts), leaner/fiber-positive options (yogurt across meal types, soups), and snack-style patterns (breads & spreads).

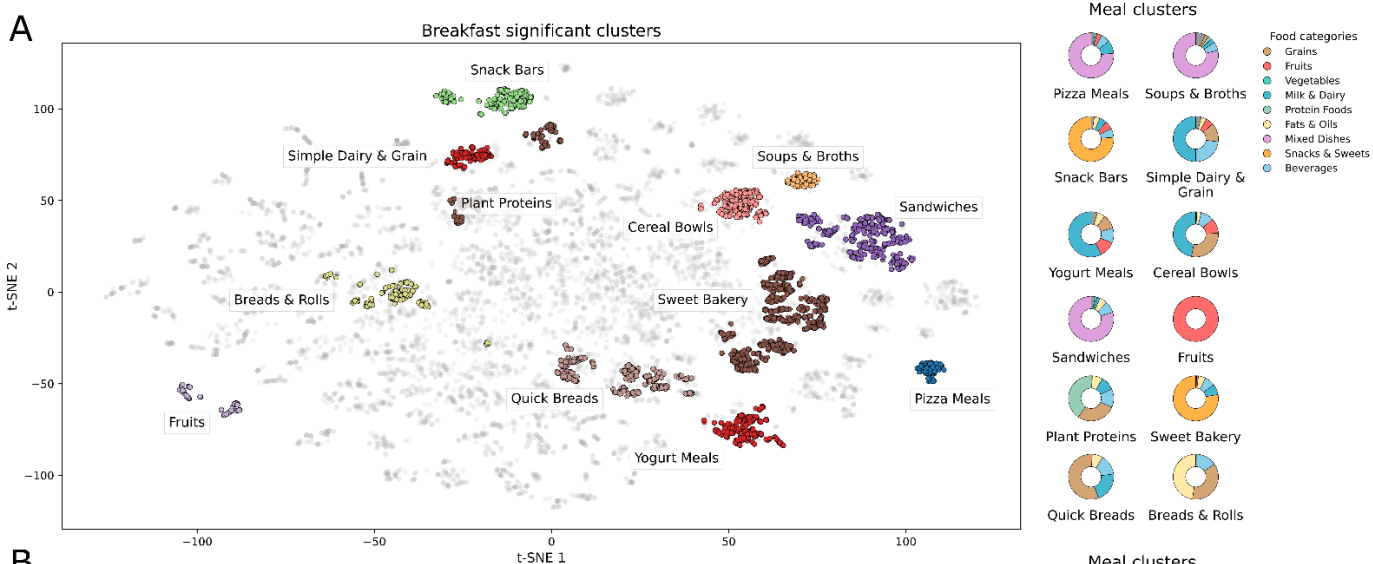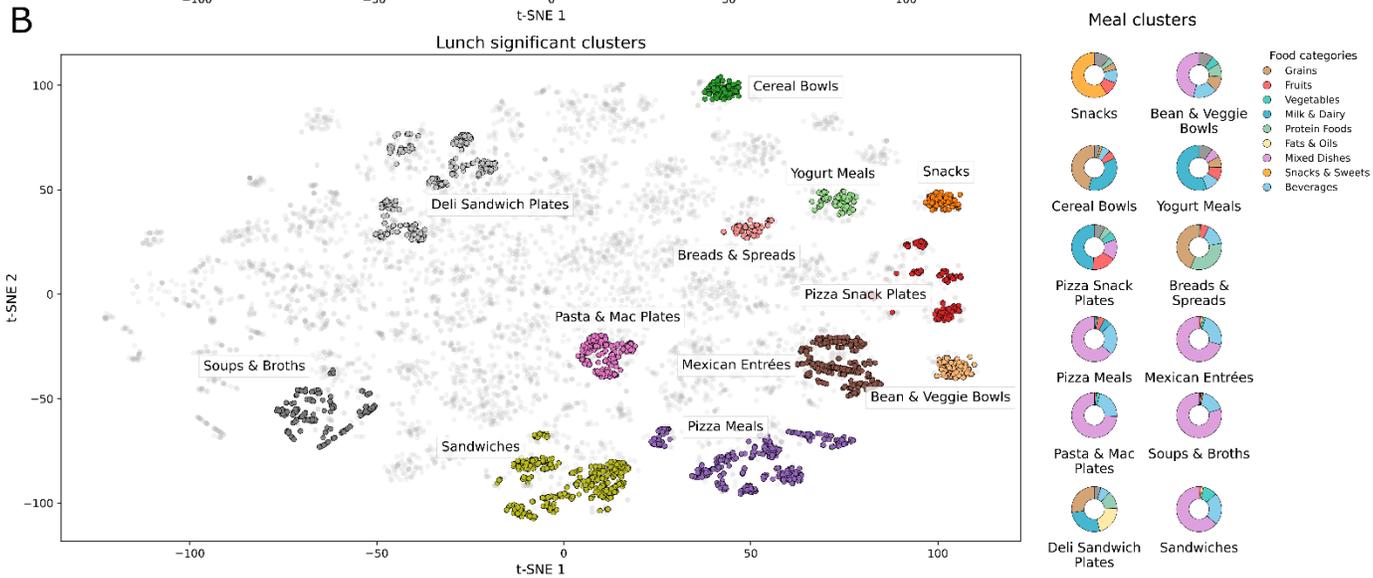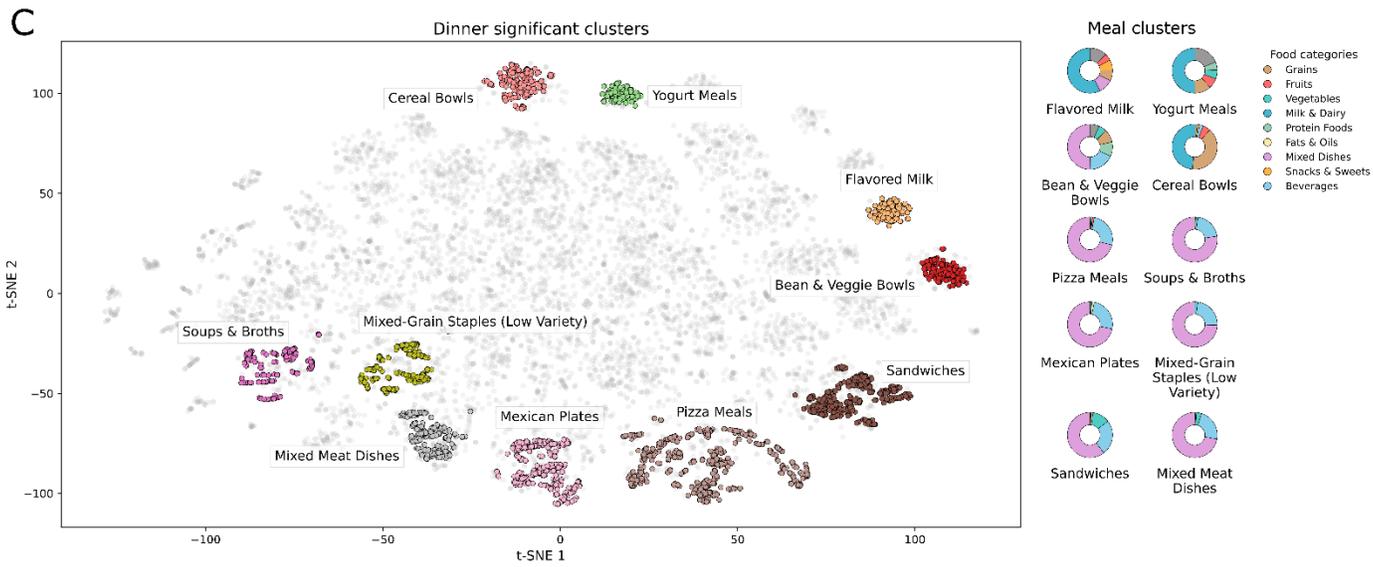

**Figure 2 | Meal archetypes across time-of-day in embedding space.** t-SNE maps of meals from the hybrid feature space (nutrients and WWEIA category grams) show interpretable, compact clusters for breakfast (top), lunch (middle), and dinner (bottom). Grey points are background meals, and colored islands are clusters that pass false discovery rate (FDR)-controlled distinctiveness tests. Labels reflect the most distinctive food-group and nutrient signals (e.g., Cereal Bowls, Sandwiches, Pizza Meals, Soups & Broths). Donut charts at right summarize each cluster's food-group composition (WWEIA main categories), illustrating the food-forward basis of the archetypes used for conditional generation and like-for-like evaluation.

**Generating realistic, uncompromised meals that improve RDI alignment.**

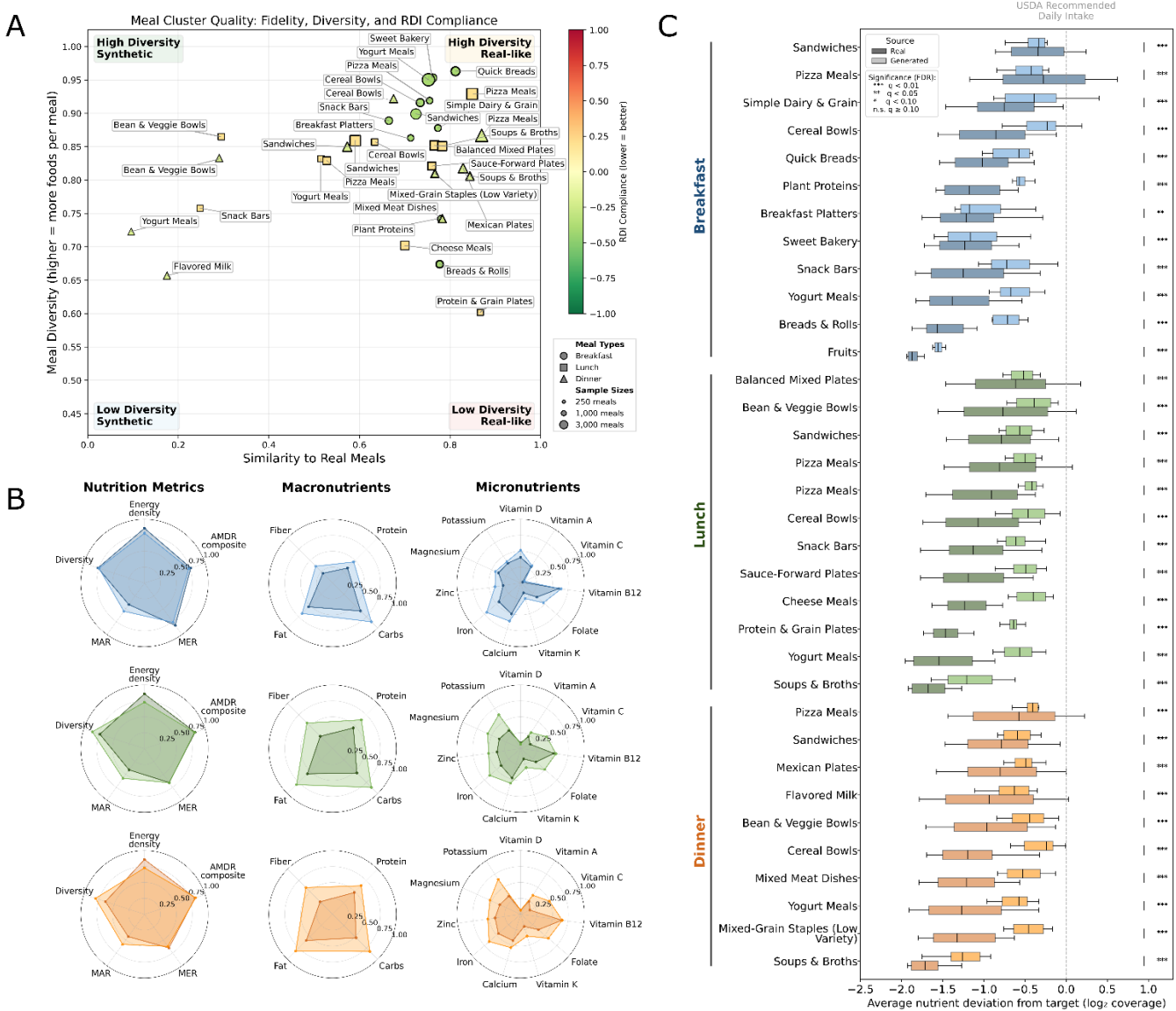

**Figure 3 | Generated meals improve nutritional targets while preserving variety.** A) Similarity–diversity map, with generated meals cluster within their archetypes yet remain widely distributed, indicating preserved variety and realism when matched by archetype and calorie band. (B) Nutritional shifts by meal type. Relative to matched real meals, MAR (adequacy) increases, AMDR compliance (balance) improves, and energy density increases; selected micronutrients (e.g., vitamin C) and protein adequacy rise, while sodium moderates. Points/boxes show bootstrap

means and 95% CIs (1,000 resamples), FDR-corrected for multiple comparisons. (C) Deviation from USDA targets. Absolute per-meal median RDI deviation drops by 47.0% overall (43.2% breakfast; 52.1% lunch; 46.0% dinner), with significant improvements across most nutrients.

We compared generated and real meals within each meal-archetype cluster using nonparametric bootstrap resampling (1,000 iterations per nutrient per cluster), counting an improvement when the 95% bootstrap interval for generated − real lay below zero. Generated meals concentrate in the high-similarity region while preserving dispersion across clusters (**Figure 3A**), while nutrition-related characteristics are substantially improving (**Figure 3B**): adequacy increases across meal types (MAR: +7.8% breakfast; +51.3% lunch; +14.4% dinner), median energy density is higher (+104.1%, +57.1%, +31.9% for breakfast, lunch, dinner), and micronutrient coverage strengthens (e.g., vitamin C +26.1%, +136.2%, +94.2%). Concomitantly, 97.1% of clusters show lower median deviation with respect to RDI recommendations for generated meals when compared to real same-cluster meals; the median percent reduction in cluster-level medians is 43.2% (breakfast), 52.1% (lunch), and 46.0% (dinner) (overall median 47.0%; **Figure 3C**). The largest improvements to RDI score are in the Dinner Cereal Bowls (−80.1% reduction to the RDI deviation), Breakfast Cereal Bowls (−73.0%), and Lunch Deli Sandwich Plates (−67.6%); the only case where deviation was increased was in the Breakfast Pizza Meals (+50.8%).

To evaluate the performance of our framework against state-of-the-art Large Language Models, we compared 3,400 meals (1,000 meals per 34 significant clusters) generated by our framework vs. similarly generated by GPT 4o (**Supplementary Material**). Results show that in all metrics except diversity, our framework over-performed GPT-4o, which is the most powerful LLMs at the time of the study, likely due to the high complexity and multi-objective nature of the task. More specifically, AMDR compliance was higher for the CVAE when compared to the LLM methods, with 18.9% vs 11.9%, respectively, of meals generated being AMDR compliant. Aggregating macronutrient composition across meals, our framework averaged 12.7% protein, 27.7% fat, and 61.6% carbohydrate, whereas GPT-4o averaged 16.8%, 39.4%, and 43.5%, and as such, GPT-4o's meals were both high-fat (>35% fat content) and low-carb (<45%). Interestingly, GPT-4o produced more diverse meals (84%) than our framework (78%), which was not one of the objectives for this study. Overall, our framework better satisfies RDI/AMDR-based adequacy and macronutrient balance in the aggregate, while GPT-4o tends to emphasize compositional diversity.

**Food substitutions can make meals 10% healthier for 32% less cost**

Can we improve existing meals both nutritionally and cost-wise with minimal change? To answer this question, we analyzed 13,656 generated meals that were one-substitution (3,858), two-substitutions (7,103), or three-substitutions (2,340) away from the total of 1,475 unique real foods. **Figure 4A** illustrates the per-meal accounting used throughout (nutrient difference and cost) for a specific meal with one and two substitutions (hops). At the knee of the cost–benefit frontier, substitutions deliver meaningful health improvements at lower cost across all hop settings. The representative knee points indicate nutrition gains of 5.7% with 19.4% savings for 1-hop, 8.1% with 30.2% savings for 2-hop, and 10.7% with 32.9% savings for 3-hop (**Figure 4B**). The composition of gains shifts systematically with hop count, with the number of hops correlating with the across-category by intra-category substitution per meal ratio, suggesting that the method has to resort to solutions from other categories to maximize benefit when multiple substitutions are allowed (**Figure 4C**). With one-hop substitutions, improvements are nutrition-led, accompanied by moderate cost relief and small adjustments to items/portions. Allowing two-hop substitutions maintains strong nutrition contribution and raises ease of adoption, indicating many options that stay close to the original plate. Under three-hop, the center of mass pivots toward savings while nutrition's share moderates; ease remains comparable. In other words, expanding flexibility widens access to deeper discounts while still improving nutrition, whereas tighter constraints emphasize nutrition-first improvements with smaller changes. Trade-off diagnostics are consistent with these patterns. Together, these results show that along the empirical frontier there are two practical operating modes, namely a local, nutrition-forward improvement with meaningful but moderate savings, to a more far-reaching, budget-forward improvements that achieve larger discounts while still moving nutrition upward.

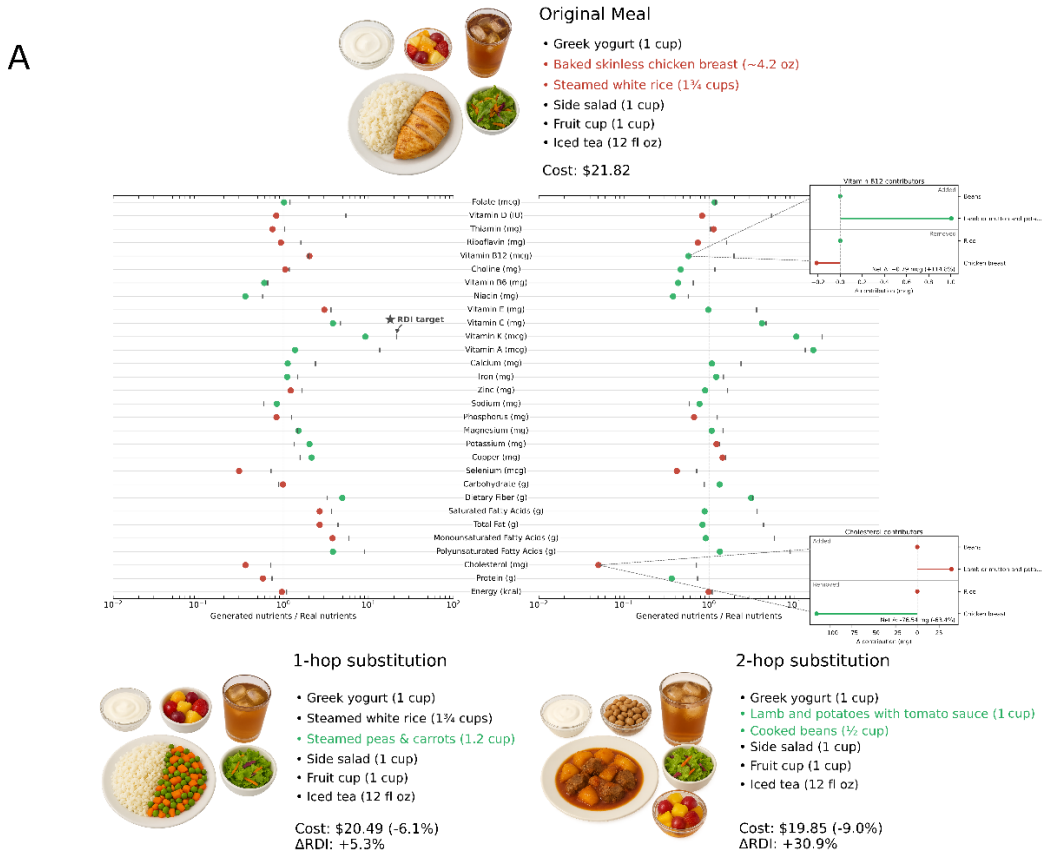

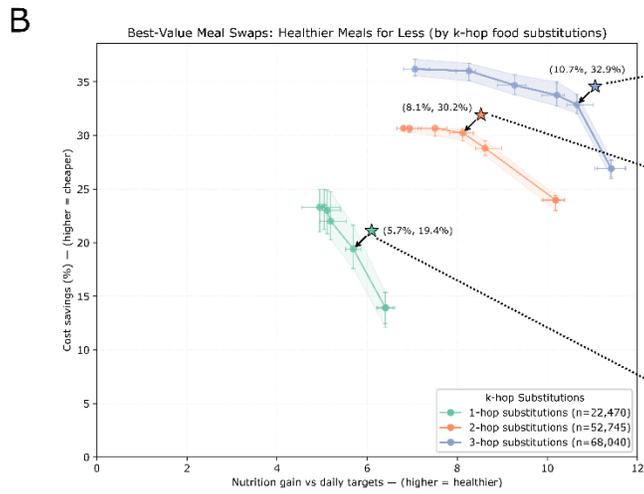

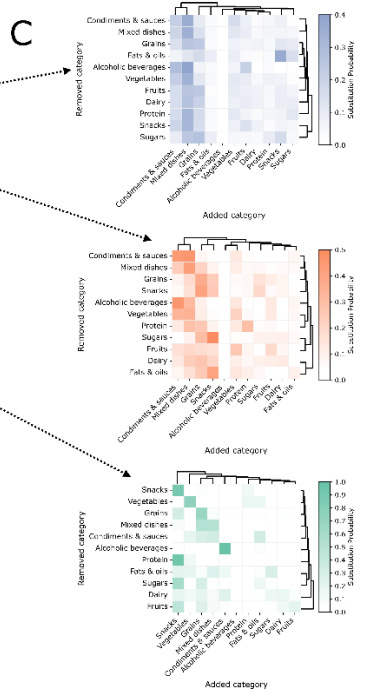

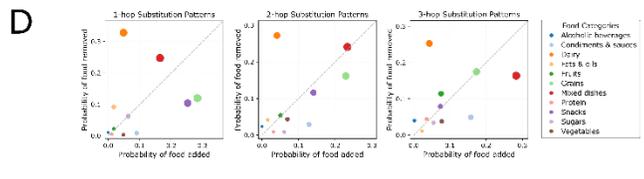

**Figure 4 | Cost–benefit of machine-guided meal substitutions**. (A) A real meal (top) and two of its generated one-hop (bottom left) and two-hop (bottom right) substitutes. The middle plot shows per-nutrient ratios (substitute/original; log scale) with the vertical line at parity. Green points to the right indicate increased progress toward daily targets; red to the left indicate decreases. Insets show benefits of 2-hop swaps include a reduction in cholesterol and an increase in vitamins. (B) Population-level cost–benefit curves summarizing, for one-, two-, and three-hop substitutions, the trade-off between health gain (percent reduction in absolute per-meal RDI deviation) and cost savings as the policy parameter θ varies. Stars indicate the selected operating points. (C) Category transition heatmaps for each starred operating point. Each cell shows the probability that the row category is swapped for the column category. Warmer colors denote more frequent directed swaps. Marginal summaries indicate net adds/removes, and clustering exposes motifs (e.g., more additions from vegetables/grains and fewer removals from high cost/low value categories), consistent with the cost–benefit gains. (D) Category-level swap scatters at the same operating points. Each point is a category, with x-axis and y-axis denoting the probability for the added and removed category, respectively. Points on the diagonal reflect in-kind replacement; right/down shifts indicate net additions, up/left indicate net removals. With higher hop counts, patterns concentrate in add-heavy, easy-to-apply substitutions, consistent with an improved cost–benefit.

## DISCUSSION

In this work, we present a meal generation and substitution framework that aims to create practical, realistic meals that optimize nutrition and cost, while imposing small changes to existing meal choices. There are several methodological advances that facilitate its improved performance. First, segmenting the meal space into meal archetypes based on nutritional and co-occurrence embeddings has been key to allowing the focused training of the machine learning engine (the conditional variational autoencoder). Similarly, separating the meal generation from portion prediction improved nutritional alignment while preserving food diversity. Without these two architecture design changes, the F1-score from reconstructing the meal inputs in the autoencoder deteriorates from 0.99 to 0.64. To quantify meal quality, we constructed multi-objective metrics to evaluate both real and generated meals based on their nutritional alignment (RDI/AMDR), realism, and cost, and found that generated meals reduced the median deviation from per-meal RDI targets by 47.0% while remaining compositionally close to observed meals. Also, key to the whole design process was to start with combinations and servings that are currently and frequently consumed, rather than ab initio plate and portion sizes. For example, to predict portion sizes that are optimized for RDI, we sample within real portion distributions per food, providing small, interpretable adjustments that correct nutrient macro- and micro- imbalances without distorting the underlying joint food combinations. Similarly, framing substitutions as a multi-objective trade-off, revealed operating points that can be tuned to user or program priorities – for example an interplay between health, affordability, or compromise to preferred choices. Head-to-head benchmarking against a LLM baseline further supports the value of embedding domain constraints directly into the generative process, and demonstrates higher standards-based performance (RDI/AMDR) than a free-form LLM approach.

Furthermore, since recommendations are derived from codified standards, the system can surface meal options as point-of-care clinical decision support, with per-nutrient deviations and a concise "why this meal" rationale (archetype match, portion adjustments, and specific swaps) to enable transparency and interpretability. In addition, expanding this framework with a human-in-the-loop step, will allow it to learn and adapt to preference signals and feedback, while keeping hard constraints, such as food allergies or clinical restrictions, in place. Similarly, guardrails can be included for populations at risk or with special nutritional needs (as for example, during pregnancy), and disease-specific recommendations, by adding extra terms and inequality constraints on the multi-objective optimization framework, and optimizing the weights for each such terms. Since one of the most important factors in whether a recommendation will be adopted both once and routinely is how difficult and different it is from the current practice, just-in-time adaptive interventions and a very well calibrated preference system that tunes in the difficulty grade of a substitution will be important. Our current work performs the first step, by minimizing and keeping track of the number of hops, however, not all substitutions are equal, and integrating a "penalty" or "effort" matrix with both a general and personalized component is likely to increase the adherence to the proposed meals by minimizing the user burden.

Future work will extend this NHANES/WWEIA-based analysis, which relies on self-reported U.S. intake, to broader cultural settings and recipe-level granularity. Prices were estimated from publicly available retail listings with category-level restaurant multipliers, and while appropriate for population-level analysis, a single, consistent pricing source would improve individual-level budgeting accuracy. Finally, we evaluate nutritional alignment and modeled costs rather than long-term adherence or clinical outcomes. Prospective, user-in-the-loop studies—potentially integrated with EHR/clinical decision support—should assess adherence, Healthy Eating Index change, and condition-specific outcomes to establish real-world impact and sustainability. These findings indicate a clear path toward actionable, guideline-consistent diet recommendations that respect preferences, budgets, and clinical constraints, and are suitable for scalable deployment across clinical and public-health settings.

## ACKNOWLEDGEMENTS

We would like to thank the members of the Tagkopoulos Lab for helpful discussions and comments. T.C. designed and performed preliminary analysis, performed all computational analysis and created figures. T.C. and I.T. contributed to the critical analysis and wrote the paper. I.T. conceived and supervised all aspects of the project.

**Funding**: This work was supported by the USDA-NIFA AI Institute for Next Generation Food Systems (AIFS), USDA-NIFA award number 2020-67021-32855, as well as the NSF HDR:TRIPODS grant CCF-1934568.

**Data Availability:** This study uses publicly available, de-identified intake data from the National Health and Nutrition Examination Survey (NHANES) "What We Eat in America" (2013–2020). Raw files can be obtained from the official repositories. Any additional data produced in this study are available from the corresponding author upon reasonable request.

**Code Availability:** All analysis and modeling code, and recreation of processed datasets that support the findings (cluster labels, evaluation outputs, and figure source data) are provided in the Supplementary Information and in our repository (https://github.com/AI-Institute-Food-Systems/Diet_Recommendation_System).

**Competing Interests:** The authors declare no competing interests.

# Translating Dietary Standards into Healthy Meals

# with Minimal Substitutions

Trevor Chan[1,2,3] and Ilias Tagkopoulos[1,2,3]*

[1]Department of Computer Science, University of California, Davis, Davis, California 95616, USA

[2]Genome Center, University of California, Davis, Davis, California, 95616, USA

[3]USDA/NSF AI Institute for Next Generation Food Systems (AIFS), University of California, Davis, Davis, California, 95616 USA

*Corresponding author (itagkopoulos@ucdavis.edu)

Supplementary materials and results

# 1. DATA ANALYSIS

## 1.1 Data Preprocessing

We processed each meal type independently from ingestion through filtering, dimensionality reduction, and modeling. We reconciled, expanded, consolidated, and renumbered USDA food codes to their most recent identifiers using the published discontinuation mappings for 2013–2020[1]. Records from earlier years were excluded due to incomplete mappings. Dropped or revised codes were retained when their usage could be consistently interpreted across waves. This yielded a dataset with 120,375 meals and 6,212 foods across all meal types[2,3]. Within each meal-type, we applied Local Outlier Factorization[4] with a neighborhood size appropriate for dense presence/gram matrices and a contamination rate of 0.3%. Scores were thresholded at the 99.7$^{th}$ percentile to remove only the most extreme composition profiles. This step removed 120 breakfast, 113 lunch, and 130 dinner meals, respectively, while preserving central structure and variance. To mitigate sparsity and improve model tractability without sacrificing nutritional signal, we merged foods into prototypes using an aggregation method guided by nutrient profiles. The method used dynamic K allocation across WWEIA subcategories with meal-type-specific parameters (α=0.10, K_max=8 per subcategory, minimum subcategory size=6). Four criteria were applied to selection and assignment:

1. Mass coverage ≥ 90% of total consumed grams to ensure prototypes represent what people actually eat,
2. Nutrient fidelity with a ≤ 7% weighted mean absolute relative error across the nutrient panel used downstream;
3. Dynamic allocation of prototype counts across WWEIA subcategories (more capacity where heterogeneity is high),

4. Assignment quality with a cosine-similarity floor of 0.70 between a food and its prototype.

This procedure reduced the number of foods by 87.5% (6,212 foods to 777 foods) while maintaining macro- and micro-nutrient distributions at the meal type level. For each meal type, we computed the binary presence of each food across meals and constructed nonparametric confidence intervals for the mean food presence in meals using 1,000 bootstrap resamples with replacement. We retained foods whose empirical mean exceeded the lower 95% confidence bound and removed foods below that bound as non-representative. Meals that became all-zero after filtering were dropped. This yielded 39,435 breakfast meals with 528 foods, 37,024 lunch meals with 627 foods, and 42,831 dinner meals with 686 foods.

**1.2 Clustering Analysis**

**1.2.1 Feature Extraction.** For each meal type (breakfast, lunch, dinner), we constructed a hybrid feature set that includes (a) nutrient totals (protein, carbohydrate, total fat, fiber, energy), macronutrient percent-of-energy ratios and balance scores, log-transformed versions of skewed nutrients, and simple macro-interaction terms; and (b) gram amounts per WWEIA category and subcategory (15 main groups and 40+ subgroups; e.g., pizza, soups, cereals, savory snacks, sweetened beverages). We also included meal-level composition indicators (e.g., item count, portion variability, calorie density, food-category diversity). The complete feature set comprised 84 features per meal, including 5 core nutritional features, 16 derived nutritional features, 53 WWEIA category gram amounts (24 main categories and 29 subcategories), 5 enhanced composition metrics, and 5 transformed features (**Supplementary Table 5**) All features were standardized by z-scoring within meal type prior to clustering.

**1.2.2 Enhanced HDBSCAN Clustering.** We used HDBSCAN[5] with a Euclidean metric and excess-of-mass (EOM) cluster selection method. Before clustering, we removed features with near-zero variance and those that were > 95% zero across meals to reduce noise. To favor generalizable clusters, we set meal-type-specific parameters: breakfast (min_cluster_size=50, min_samples=25, alpha=1.0, cluster_selection_epsilon=0.1), lunch (min_cluster_size=40, min_samples=20, alpha=1.0, cluster_selection_epsilon=0.08), and dinner (min_cluster_size=35, min_samples=18, alpha=1.0, cluster_selection_epsilon=0.06). These parameters were adjusted to reflect distributional differences (breakfast is typically more diverse while dinner is more homogeneous). After clustering, small clusters were merged into the most similar larger cluster using centroid similarity (merge when cosine similarity exceeded 0.7; otherwise assign to the nearest large cluster by Euclidean distance). This post-processing step increased coverage while preserving separability of dominant meal patterns.

**1.2.3 Statistical Validation.** For each cluster, we compared feature distributions against the complement set using a two-part hurdle approach: a prevalence test (non-zero vs. zero; Fisher's exact[6] where applicable) and an intensity test on non-zeros (two-sided Mann–Whitney[7]). For each feature, we retained the p-value, then applied Benjamini–Hochberg FDR control at $q \leq 0.01$[8] across all cluster–feature pairs. We required a minimum absolute mean difference of 0.15 to declare significance and flagged distinctive features at $|\Delta| \geq 0.20$. We report Cohen's d[9] to summarize effect sizes. Clusters dominated by beverages or exhibiting ultra-sparse compositions were deprioritized during interpretation to maintain food-forward, practically actionable archetypes. Cluster names were manually annotated by observing significant descriptors of each cluster. These clusters serve as conditioning variables in the generative model and as strata for evaluation.

## 2. MEAL GENERATION PIPELINE

### 2.1 Food Composition Generation

We use a conditional VAE (CVAE)[10] to predict the presence of foods, jointly conditioned on cluster and meal type. The encoder is an MLP (ReLU, 0.1 dropout) producing a 64-dimensional Gaussian latent z. The decoder is a presence head with FiLM modulation[11]: 3 Dense(512, GELU[12]) blocks, each FiLM-conditioned by the concatenated cluster and meal embeddings (32-dim each). Two structural priors are installed: (i) a learned pair-specific prior that boosts likely foods per (meal, cluster), and (ii) a hard allowed-foods gate that zeroes disallowed items (also applied as a hard mask at inference). Training uses Adam[13] (learning rate of 0.0005) and a batch size of 128 over 50 epochs. The loss is weighted binary cross-entropy with dynamic positive-class weighting plus KL with free-bits[14] and an annealed β schedule[15] (linear warmup to 0.01, then light triangular cycling).

### 2.2 Portion Assignment

For portion assignment, we employed an RDI-per-kilocalorie optimization. For each generated meal from the CVAE, foods with predicted presence probabilities above a threshold ($p \geq 0.02$) are retained up to a maximum of 12 items. From the daily RDI vector $R$, we compute per-kcal nutrient densities

$$t_k = \frac{R_k}{R_{\text{energy}}}$$

where $R_{\text{energy}} = 2{,}000$ kcal. Each meal $m$ receives a fractional energy allocation $f_{meal\ type} \in \{breakfast: 0.25, lunch: 0.35, dinner: 0.40\}$, giving meal-scaled targets

$$r_k = t_k\, f_m\, R_{\text{energy}}.$$

Vitamin D units are auto-converted (IU to µg) before normalization. We let $A \in \mathbb{R}^{n \times m}$ denote per-gram nutrient densities and $x \in \mathbb{R}^n$ the portions(g) of $n$ selected foods. The optimization minimizes asymmetric log₂ deviations from meal-scaled targets:

$$\min_{x \geq 0} [\sum_k w_k (\log_2 \frac{(Ax)_k}{r_k})_+^2 + \sum_{k'} v_{k'} (\log_2 \frac{r_{k'}}{(Ax)_{k'}})_+^2],$$

where $w_k$ and $v_k$ weight under- and over-consumption penalties (defaults listed in **Supplementary Table 10**). High-priority adequacy nutrients (energy, protein, fiber, potassium) receive greater weights (2.0), while moderation nutrients (sodium, added sugars, saturated fat) receive stronger penalties (3 to 4). Portions are constrained by:

- Total grams: ≤ 900 g per meal; soft downscaling if exceeded
- Beverages: ≤ 25% kcal of meal energy, with per-meal caps 300g/350g/350g (breakfast/lunch/dinner)
- Component caps: Sugars ≤ 12g, Fats/Oils ≤ 20g, Condiments/Sauces ≤ 20g, Snacks/Sweets ≤ 60g
- Per-item bounds: Solid foods ≤ 300 g each; minimum solid items = 2/3/3 (breakfast/lunch/dinner)

Optimization proceeds until the meal energy target is matched. Nutrient totals are re-projected if group caps bind. The final portion vector $x^*$ defines a nutritionally coherent meal in grams. Derived nutrients are reported as $(x^*/100)^\top A_{100g}$.

## 2.3 Comparison against state-of-the-art LLM

We evaluated our CVAE against a state-of-the-art large language model, GPT-4o for meal recommendation. An example of a prompt that was used to generate breakfast meals is the following:

> **Prompt Example: Breakfast Meal Generation**
>
> **Instruction:** Create a diverse breakfast meal.
>
> **NUTRITIONAL TARGETS**
>
> - Energy: 500 kcal ± 5%
> - Protein 20.0 g; Fiber 7.5 g
> - Sodium 450.0 mg; Added sugars 10 % energy
> - Saturated fat (SFA) 10 % energy
> - Beverage energy 25 %
>
> **STRUCTURAL CONSTRAINTS**
>
> - 2–4 items total; 1 beverage; 2 solids
> - Max grams: solids 300 g, beverages 500 g
>
> **PORTION GUIDELINES**
>
> - Vegetables 30–200 g; Fruits 50–300 g
> - Grains 15–100 g; Protein 15–150 g
> - Dairy 30–500 g; Beverages 120–500 g
>
> **CATALOG (code → name)**
> ```
> { "11100000": "Milk, NFS", "11320000": "Soy milk", "11436000":
> "Yogurt, liquid",
>  "11513550": "Chocolate milk (made from syrup with reduced fat
> milk)", ...}
> ```

where we limited GPT-4o's context to include only the inputs provided to our generator, i.e. only foods considered within cluster were used by GPT-4o in its meal generation process, as well as the constraints, i.e. nutritional targets, provided to our portion assignment module.

## 3. FOOD SUBSTITUTION

**3.1 Pricing Data.** We price meals using a portion-based restaurant model derived from public menu listings and industry guides. For each item $j$, we specify a grams-per-portion $g_j^{(\text{portion})}$ and a price-per-portion $p_j^{(\text{portion})}$. Given the meal's actual grams $w_j$, we compute a capped portion multiplier

$$m_j = \min\left(\frac{w_j}{g_j^{(\text{portion})}}, c_j\right),$$

where $c_j$ is a category-dependent cap that limits how many portions of an item can be charged within a single meal. The meal cost is then

$$C = \sum_j m_j \, p_j^{(\text{portion})} + \omega,$$

with a configurable overhead parameter $\omega$, in experiments it was set to $2. Certain categories were capped, for example soups and fruit-salad $c_j = 1.0$; sides $c_j \leq 2.0$; main entrees and mixed dishes $c_j \leq 1.5$. We used a GPT-5 deep-researched grocery pricing matrix per-100 grams of food for any uncategorized foods. We also applied cross-item caps (e.g., at most one soup bowl and one fruit-salad serving per meal) for added meal practicality.

**3.2 Evaluation.** We treat food substitutions in a meal as a constrained nearest-neighbor problem[16] over real meals. From the real meal dataset, we analyzed 13,656 meals for which a generated substitute could be found, drawn from 1,475 unique real foods (Breakfast 455, Lunch 493, Dinner 527; rows by type: Breakfast 4,275, Lunch 4,625, Dinner 4,756). For each generated meal, we form a candidate pool by retrieving the k-nearest real meals using a similarity that mixes presence Jaccard[17] and portion-cosine[18] (weight 0.7/0.3), under energy and item count. Similarity combines item overlap and portion composition; we require comparable energy (±5%) and similar item count (±1). For each candidate pair we compute edits and define $k_{\text{sub}} = \max\{\#\text{added}, \#\text{removed}\}$. Analyses are run separately at $k_{\text{sub}} \in \{1,2,3\}$ by enforcing exact $k_{\text{sub}}$ at selection time. Subsequently, we define

- **Nutrition gain** $H_i$ as the positive change in percent of daily targets achieved. We use $H_i = \max(0, \Delta RDI_i)$, so that negative changes do not contribute to "gain."

- **Cost saving** $S_i$ as the positive percent reduction in price,

$$S_i = \max(0, \frac{Cost_{real} - Cost_{substitution}}{Cost_{real}} \times 100\%).$$

- **Swap effort** $E_i \in [0,1]$ as the behavioral burden of the substitution, if taken. We combine (i) portion shift - the $L_1$ change in ingredient weights, normalized by the meal's total weight - and (ii) composition change - the larger of "added" or "removed" ingredients. Effort is $k_{\text{sub}}$-aware - we normalize portion and composition changes against k-based baselines, and set $E_i = \alpha E_{portion} + (1 - \alpha) E_{composition}$, with $\alpha = 0.5$.

**3.3 Winner Selection.** For a trade-off preference $\theta \geq 0$, we map to a health weight $w = \frac{\theta}{1+\theta} \in [0,1]$. Each candidate receives a value

$$V_i(\theta) = wH_i + (1-w)S_i - wCI_i,$$

where

- $H_i = \max(0, \Delta\text{RDI}_i)$ is the health gain in percentage points (reduction in mean absolute deviation from per-meal RDI targets).

- $S_i = \max\left(0, \frac{\text{Cost}_{\text{real}} - \text{Cost}_{\text{sub}}}{\text{Cost}_{\text{real}}} \times 100\%\right)$ is cost saving (non-negative by construction).

- $CI_i = \max\left(0, \frac{\text{Cost}_{\text{sub}} - \text{Cost}_{\text{real}}}{\text{Cost}_{\text{real}}} \times 100\%\right)$ is the cost-increase penalty (zero when the substitute is cheaper or equal).

We omit candidates that add or remove items from beverages or other main categories; pure portion-only edits are allowed but treated as within-category. Substitution winning candidates with the highest $V_i(\theta)$ are chosen in two stages:

1. Within-category stage. If any candidates keep the WWEIA main category matched between added and removed items (or only change portions), we shortlist them and pick the top by the value score below.

2. Cross-category challenge. A cross-category candidate may overtake the provisional within-category winner only if it is clearly better. If it adds any mixed dishes, it must exceed the within-category score by both a relative margin $\alpha = 0.25$ and an absolute buffer $\beta = 1.5$; otherwise a $\geq 20\%$ relative uplift is required. If no within-category options exist, we select from the pooled set.

Ties are broken by (i) smaller portion-shift percentage, then (ii) larger health increase, then (iii) smaller cost savings. We ignore candidates with negative health change and drop negative overall scores when considering substitution candidates.

**3.4 Cost–benefit aggregation.** Sweeping $\theta$ across a grid spanning health-tilted to cost-tilted regimes, we select per-meal winners with $V_i(\theta)$, report the median nutrition gain $\tilde{H}(\theta)$ and median savings $\tilde{S}(\theta)$ across meals, and form 95% bootstrap CIs over meals (1,000 resamples).

# 4. ALGORITHMIC IMPLEMENTATION

Putting everything together, our framework, executes the following workflow.

---

**Algorithm 1: Framework for Standard-Aligned Meal Generation and Substitution**

**Input:** NHANES/WWEIA (2013–2020) meals $\{M_i\}$ with food codes and nutrients
**Output:** RDI-compliant generated meals and cost-effective substitutions

1 **Stage 1: Data Preprocessing** `Harmonize, clean, and filter raw records`
2 **begin**
3     Harmonize USDA food codes across survey waves via discontinuation maps
4     Remove incomplete or outdated records; drop outliers via Local Outlier Factorization (0.3%)
5     Aggregate foods into nutrient-based prototypes preserving $\geq 90\%$ mass coverage
6     Retain representative foods whose bootstrap lower CI for presence $> 0$

7 **Stage 2: Meal Archetype Clustering** `Identify interpretable time-of-day archetypes`
8 **begin**
9     Construct hybrid feature matrix $\mathbf{X}$: nutrient totals, ratios, and WWEIA category grams
10    Z-score standardize features within meal type (B/L/D)
11    Apply HDBSCAN with meal-type-specific parameters; merge small clusters via cosine similarity $> 0.7$
12    Validate features via FDR-controlled tests ($q \leq 0.01$) and label interpretable clusters $\mathcal{A}_m$

13 **Stage 3: Conditional Meal Generation (CVAE)** `Generate realistic food combinations`
14 **begin**
15    Encoder $E_\phi(x, c_m, c_t) \to \mathcal{N}(\mu_z, \sigma_z)$; Decoder $D_\theta(z, c_m, c_t)$ predicts presence probs
16    Apply FiLM conditioning on cluster ($c_m$) and meal type ($c_t$)
17    Train with weighted BCE + annealed KL loss:

$$\mathcal{L} = \mathbb{E}_{q_\phi(z|x,c)}[-\log p_\theta(x|z,c)] + \beta D_{KL}(q_\phi || p(z))$$

      Optimize via Adam (lr = 5e-4), batch = 128, 50 epochs
18    Retain foods with $p_i \geq 0.02$, up to 12 per meal

19 **Stage 4: RDI-per-kcal Portion Assignment** `Convert combinations to balanced portions`
20 **begin**
21    Compute per-kcal nutrient targets $t_k = \frac{RDI_k}{E_{daily}}$
22    Assign energy fractions $(0.25, 0.35, 0.40)$ for (B/L/D)
23    Solve for portions $g_i$ minimizing asymmetric log-deviation:

$$\min_g \sum_k w_k \left| \log_2 \left( \frac{\sum_i g_i n_{ik}}{t_k} \right) \right|$$

      Subject to: total grams $\leq 900$, beverages $\leq 25\%$ kcal, solids $\leq 300$g, and category caps
24    Reproject totals if caps bind to restore energy targets

25 **Stage 5: Substitution Optimization** `Find practical cost{health trade-offs`
26 **begin**
27    For each generated meal $M_g$, retrieve $k$ nearest real meals $M_r$ via Jaccard–cosine similarity
28    Ensure comparable energy ($\pm 5\%$) and item count ($\pm 1$)
29    For each $(M_g, M_r)$ pair compute:

$$\Delta_{nut} = \max(0, RDI_{gain}), \quad \Delta_{cost} = \max\left(0, \frac{Cost_r - Cost_g}{Cost_r}\right)$$

      Compute swap effort $\mathcal{E}$ from portion and composition changes
30    Select winner maximizing:

$$V = \lambda \Delta_{nut} + (1 - \lambda) \Delta_{cost} - \alpha \mathcal{E}$$

      Prefer within-category swaps; allow cross-category if $\geq 20\%$ relative improvement
31    Aggregate median nutrition and cost gains over 1,000 bootstrap resamples

32 **Return:** Final meals with optimized nutrition, realistic structure, and minimal-effort substitutions.

**SUPPLEMENTARY TABLES**

|  | Number of Records | Number of Respondents | Number of Meals | | |
|---|---|---|---|---|---|
|  |  |  | Breakfast | Lunch | Dinner |
| NHANES 2013-2014 Day 1 | 131,394 | 8,661 | 21,331 | | |
|  |  |  | 7,019 | 6,594 | 7,718 |
| NHANES 2013-2014 Day 2 | 112,578 | 7,573 | 19,250 | | |
|  |  |  | 6,367 | 6,074 | 6,809 |
| NHANES 2015-2016 Day 1 | 121,481 | 8,505 | 20,501 | | |
|  |  |  | 6,789 | 6,298 | 7,414 |
| NHANES 2015-2016 Day 2 | 100,680 | 7,027 | 17,539 | | |
|  |  |  | 5,827 | 5,538 | 6,174 |
| NHANES 2017-2020 Day 1 | 183,910 | 12,632 | 30,242 | | |
|  |  |  | 9,914 | 9,210 | 11,118 |
| NHANES 2017-2020 Day 2 | 149,495 | 10,830 | 26,628 | | |
|  |  |  | 8,865 | 8,244 | 9,519 |
| **Total** | **799,538** | **55,228** | **135,491** | | |
|  |  |  | **44,781** | **41,958** | **48,752** |

**Supplementary Table 1** | Table showing the number of records, respondents, and meals per NHANES survey from 2013-2020.

| Term | Definition |
|---|---|
| Nutrients | the atomic units of meals e.g. vitamin C, Iron |
| Ingredients | composed of multiple nutrients, has a USDA ingredient code e.g. rice, seaweed |
| Food | composed of multiple ingredients, has a USDA food code e.g. sushi, sake |
| Meal | composed of multiple foods e.g. sushi with sake |
| Real Meal | meal that is consumed by real people (ground truth) |
| Practical Meal | meal that resembles a meal that is consumed by the population |
| Time-of-day | the eating occasion of a given meal e.g. breakfast, lunch, dinner |
| Healthy | as nutritionally close to USDA nutritional guidelines as possible |
| Recipe | ingredients with a list of instructions on how to cook them to become foods |

**Supplementary Table 2** | Definitions of meal composition framework

| Discontinued Food Code | Description |
|---|---|
| 1 – Dropped | Products no longer on the market; codes rarely used in the survey; items better coded using individual components as a combination. The specific products represented by some deleted codes have changed and no longer represent the food/beverage as described; however, similar products may be present within new/existing codes. |
| 2 – Expanded | New codes were created and the original code was discontinued. Modifications eliminated - products expanded to include presence/type of fat; codes expanded to designate source (restaurant, fast food, school); codes expanded by different ingredients. Codes representing more than one product or variety were expanded to individual codes. |
| 3 – Consolidated | Multiple codes now captured under a new/existing single code. The original codes were discontinued. |
| 4 – Renumbered | A food code was assigned a different 8-digit number yet represents the same product. Main food description may have been revised. |
| 5 – Revised | Codes received extensive revisions, including expansion and consolidation. |

**Supplementary Table 3** | FNDDS discontinued food codes.

|  | Number of Meals | | | Number of Unique Foods | | |
| --- | --- | --- | --- | --- | --- | --- |
|  | Breakfast | Lunch | Dinner | Breakfast | Lunch | Dinner |
| Raw Dataset | 135,491 | | | 7,679 | | |
|  | 44,781 | 41,958 | 48,752 | 4,346 | 6,063 | 6,633 |
| Food code standardization and filtration of unusable data | 134,457 | | | 6,599 | | |
|  | 44,440 | 41,628 | 48,389 | 3,881 | 5,361 | 5,804 |
| Outlier Filtration | 134,052 | | | 6,599 | | |
|  | 44,306 | 41,503 | 48,243 | 3,881 | 5,361 | 5,804 |
| Bootstrap Confidence Intervals | 134,052 | | | 6,591 | | |
|  | 44,306 | 41,503 | 48,243 | 3,872 | 5,356 | 5,788 |
| Food composition Filtration | 65,202 | | | 2018 | | |
|  | 25,675 | 19,674 | 19,853 | 378 | 794 | 846 |

**Supplementary Table 4 | The data visualization of the dataset size in terms of meals and their corresponding foods at each preprocessing step.** Table showing the preprocessing steps and resulting data dimensionality

| Feature Category | Count | Reformatted Feature Names | Description |
|---|---|---|---|
| **Core Nutritional Features** | 5 | Protein (g), Carbohydrate (g), Fat (g), Fiber (g), Energy (kcal) | Basic macronutrients and energy |
| **Derived Nutritional Features** | 16 | Protein Ratio, Carbohydrate Ratio, Fat Ratio; Protein Level, Carbohydrate Level, Fat Level, Energy Level; Protein–Carbohydrate Balance, Protein–Fat Balance, Carbohydrate–Fat Balance; Meal Balance Score, Nutritional Balance; Grain Ratio, Vegetable Ratio, Fruit Ratio, Dairy Ratio | Macronutrient and meal balance indicators |
| **WWEIA Main Categories** | 24 | Milk/Dairy, Mixed Dishes, Grains, Snacks/Sweets, Fruits, Vegetables, Beverages, Alcoholic Beverages, Water, Condiments/Sauces, Sugars, Baby Foods, Other; Milk, Flavored Milk, Dairy Drinks, Cooked Grains, Savory Snacks, Diet Beverages, Sweetened Beverages, Plain Water, Flavored Water, Baby Beverages, Human Milk | Main WWEIA categories |
| **WWEIA Subcategories** | 29 | Protein Foods, Fats/Oils, Cheese, Yogurt, Meats, Poultry, Seafood, Eggs, Cured Meats, Plant Proteins; Mixed Meat Dishes, Mixed Bean Dishes, Mixed Grain Dishes, Asian Dishes, Mexican Dishes, Pizza, Sandwiches, Soups; Breads/Rolls, Quick Breads, Cereals, Crackers, Snack Bars, Sweet Bakery, Candy, Desserts; Juice, Coffee/Tea, Infant Formulas | Protein, mixed dish, grain/snack, and beverage subcategories |
| **Composition Features** | 5 | Macronutrient Diversity, Food Category Diversity, Ingredient Count, Portion Variability, Calorie Density | Meal-level composition indicators |
| **Transformed Features** | 5 | log Protein (g), log Carbohydrate (g), log Fat (g), log Energy (kcal), log Fiber (g) | Log-transformed nutritional features |

**Supplementary Table 5 | Summary of engineered meal-level feature categories used in the analysis.** This table lists all features derived from USDA and WWEIA mappings, grouped into six major categories. Core Nutritional Features capture basic macronutrient quantities and energy values. Derived Nutritional Features include macronutrient ratios, balance scores, and food group proportions. WWEIA Main and Subcategories represent hierarchical food groupings spanning milk, grain, fruit, vegetable, protein, and beverage domains. Composition Features quantify meal-level diversity, portion variability, and calorie density, while Transformed Features provide log-scaled nutrient representations for normalized statistical modeling.

| Meal Type | Cluster Name | Number of Meals | Number of Foods | Top Categories | Max Absolute Cohen's d |
|---|---|---|---|---|---|
| **Breakfast** | Pizza Meals | 238 | 59 | Pizza; Mixed dishes; Flavored milk; Sweetened beverages; Diet beverages | 43.32 |
| | Soups & Broths | 223 | 59 | Soups; Mixed dishes; Plain water; Condiments sauces; Flavored milk | 14.19 |
| | Snack Bars | 549 | 56 | Snack bars; Snacks sweets; Plain water; Yogurt; Mixed bean dishes | 15.57 |
| | Simple Dairy & Grain | 277 | 49 | Other; Milk; Milk dairy; Plain water; Mixed bean dishes | 8.25 |
| | Yogurt Meals | 590 | 51 | Yogurt; Milk dairy; Fruits; Plain water; Cereals | 9.77 |
| | Cereal Bowls | 508 | 50 | Dairy drinks; Cereals; Milk dairy; Grains; Fruits | 8.23 |
| | Sandwiches | 1122 | 55 | Sandwiches; Mixed dishes; Sweetened beverages; Beverages; Coffee tea | 8.66 |
| | Fruits | 301 | 12 | Fruits; Mixed bean dishes; Flavored water; Savory snacks | 12.89 |
| | Plant Proteins | 213 | 41 | Plant proteins; Breads rolls; Protein foods; Grains; Coffee tea | 4.36 |
| | Sweet Bakery | 1871 | 45 | Sweet bakery; Snacks sweets; Coffee tea; Mixed bean dishes | 4.22 |
| | Quick Breads | 711 | 38 | Quick breads; Grains; Sugars; Milk; Mixed bean dishes | 3.04 |
| | Breads & Rolls | 396 | 35 | Breads rolls; Fats oils; Coffee tea; Sugars; Grains | 1.92 |
| **Lunch** | Snacks | 247 | 69 | Snack bars; Snacks sweets; Fruits; Crackers; Plain water | 16.53 |
| | Bean & Veggie Bowls | 278 | 66 | Mixed bean dishes; Mixed dishes; Plain water; Meats; Cooked grains | 15.13 |
| | Cereal Bowls | 271 | 67 | Cereals; Milk; Grains; Milk dairy; Sugars | 11.85 |
| | Yogurt Meals | 218 | 61 | Yogurt; Milk dairy; Plain water; Cooked grains; Fruits | 8.85 |
| | Pizza Snack Plates | 465 | 51 | Flavored milk; Milk dairy; Pizza; Poultry; Fruits | 5.6 |
| | Breads & Spreads | 210 | 40 | Sugars; Plant proteins; Breads rolls; Grains; Protein foods | 4.28 |
| | Pizza Meals | 1242 | 49 | Pizza; Mixed dishes; Sweetened beverages; Milk; Beverages | 5.46 |
| | Mexican Entrées | 767 | 41 | Mexican dishes; Mixed dishes; Sweetened beverages; Plain water; Beverages | 4.26 |
| | Pasta & Mac Plates | 527 | 38 | Mixed grain dishes; Mixed dishes; Plain water | 3.91 |
| | Soups & Broths | 761 | 38 | Soups; Mixed dishes; Plain water | 4.34 |
| **Lunch Dinner** | Deli Sandwich Plates | 703 | 44 | Cheese; Breads rolls; Cured meats; Grains; Milk dairy | 2.05 |
| | Sandwiches | 1124 | 38 | Sandwiches; Mixed dishes; Sweetened beverages; Beverages; Condiments sauces | 4.02 |
| | Flavored Milk | 352 | 70 | Flavored milk; Milk dairy; Sweet bakery; Pizza; Desserts | 24.54 |
| | Yogurt Meals | 304 | 72 | Yogurt; Milk dairy; Cooked grains; Plant proteins; Plain water | 17.4 |
| | Bean & Veggie Bowls | 369 | 64 | Mixed bean dishes; Mixed dishes; Plain water; Beverages; Mixed meat dishes | 10.8 |
| | Cereal Bowls | 551 | 61 | Cereals; Milk; Milk dairy; Grains; Sugars | 9.54 |
| | Sandwiches | 884 | 44 | Sandwiches; Mixed dishes; Sweetened beverages; Beverages; Condiments sauces | 4.93 |
| | Pizza Meals | 1715 | 43 | Pizza; Mixed dishes; Sweetened beverages; Snack bars | 4.99 |
| | Soups & Broths | 578 | 36 | Soups; Mixed dishes; Plain water; Snack bars | 3.82 |
| | Mexican Plates | 850 | 37 | Mexican dishes; Mixed dishes; Sweetened beverages; Plain water; Snack bars | 3.72 |
| | Mixed Meat Dishes | 537 | 38 | Mixed meat dishes; Mixed dishes; Plain water; Snack bars | 2.92 |
| | Mixed-Grain Staples (Low Variety) | 504 | 31 | Mixed grain dishes; Mixed dishes; Plain water; Snack bars | 2.59 |

**Supplementary Table 6 | Meal distributions per cluster.** The number of meals and unique foods per significant cluster, and the top contributing food categories. These meals form the basis for the training of the meal generation model (CVAE), and upon generation the same number of meals per cluster was used for evaluation.

| Metric | Breakfast | Lunch | Dinner |
|---|---|---|---|
| $F_1$ (micro) | 0.9967 ± 0.0008 | 0.9969 ± 0.0007 | 0.9975 ± 0.0004 |
| $F_1$ (macro) | 0.8337 ± 0.0396 | 0.8417 ± 0.0328 | 0.8486 ± 0.0342 |
| Precision (micro) | 0.9967 ± 0.0008 | 0.9969 ± 0.0007 | 0.9975 ± 0.0004 |
| Precision (macro) | 0.8393 ± 0.0441 | 0.8564 ± 0.0410 | 0.8877 ± 0.0149 |
| Recall (micro) | 0.9967 ± 0.0008 | 0.9969 ± 0.0007 | 0.9975 ± 0.0004 |
| Recall (macro) | 0.8291 ± 0.0397 | 0.8287 ± 0.0278 | 0.8189 ± 0.0473 |
| AUROC | 0.9844 ± 0.0022 | 0.9834 ± 0.0024 | 0.9740 ± 0.0023 |
| AUPRC | 0.7317 ± 0.0893 | 0.7447 ± 0.0764 | 0.7532 ± 0.0734 |
| Brier score | 0.00235 ± 0.00047 | 0.00223 ± 0.00040 | 0.00195 ± 0.00035 |
| Count $R^2$ | 0.7689 ± 0.0179 | 0.8562 ± 0.0289 | 0.7593 ± 0.0441 |
| Count Bias | 0.0763 ± 0.0077 | 0.0746 ± 0.0109 | 0.1198 ± 0.0204 |
| True count (mean) | 2.530 ± 0.014 | 2.480 ± 0.025 | 2.254 ± 0.016 |
| Pred count (mean) | 2.606 ± 0.007 | 2.555 ± 0.027 | 2.374 ± 0.029 |

**Supplementary Table 7 | Performance of the presence-only CVAE model under five-fold cross-validation, reported as mean ± standard deviation across folds for each meal type**. Metrics include classification fidelity ($F_1$, precision, recall), calibration (AUROC, AUPRC, Brier), and quantitative agreement in predicted food counts ($R^2$, Bias). Presence thresholds denote optimal probability cut-offs per fold, while "True" and "Pred count" indicate average numbers of foods per meal in ground-truth and model predictions, respectively.

| Meal Type | Energy Fraction | Energy Target (kcal) | Description |
|---|---|---|---|
| Breakfast | 0.25 | 500 | Represents 25% of daily 2000 kcal target |
| Lunch | 0.35 | 700 | Represents 35% of daily 2000 kcal target |
| Dinner | 0.4 | 800 | Represents 40% of daily 2000 kcal target |

**Supplementary Table 8 | Meal energy-fraction assignments used for RDI-per-kcal portioning.** The fractional distribution of daily energy across breakfast, lunch, and dinner under a 2000 kcal pattern (0.25 / 0.35 / 0.40). These fractions are applied to both total caloric and per-kcal nutrient targets in the RDI-per-kcal optimization. The allocation reflects typical USDA dietary guidance and ensures that each generated meal represents its proportional share of daily intake.

| Category / Subgroup | Cap (g) | Condition / Rule | Rationale |
|---|---|---|---|
| **Total meal weight** | ≤ 900 | per meal | Limit total grams for realistic plate size |
| **Beverages (total)** | ≤ 25 % kcal or ≤ 300 / 350 / 350 g (B/L/D) | per meal | Avoid excessive beverage volumes |
| **Added sugars** | ≤ 12 g | absolute | Approx. 10 % kcal soft cap (per 2000 kcal diet) |
| **Fats / Oils** | ≤ 20 g | absolute | Prevent over-representation of oils/fats |
| **Condiments / Sauces** | ≤ 20 g | absolute | Maintain realistic side condiment levels |
| **Snacks / Sweets** | ≤ 60 g | absolute | Limit sweet or discretionary foods |
| **Solid food items (per item)** | ≤ 300 g | per item | Avoid unrealistic single-item servings |
| **Minimum # solid items** | 2 (B) / 3 (L) / 3 (D) | per meal | Ensure variety and realism |
| **Fluid dairy count as beverage** | – | inclusion rule | Included in beverage cap |

**Supplementary Table 9 | Group- and per-item gram caps enforced during RDI-per-kcal optimization.** Gram-level caps and qualitative rules used to bound meal size, beverage volume, and individual component weights. These constraints prevent unrealistic portions while preserving dietary realism and nutrient feasibility. Beverage energy is limited to ≤ 25 % of meal kcal (300 / 350 / 350 g for breakfast, lunch and dinner), total meal mass is capped at ≤ 900 g, and per-item solids at ≤ 300 g. Additional limits on sugars, fats/oils, condiments, and snacks constrain discretionary energy. All caps are enforced softly via proportional rescaling when totals exceed limits.

| Nutrient | Default RDI | RDI-per-kcal | Weight (Under) $w_k$ | Weight (Over) $v_k$ | Constraint Type |
|---|---|---|---|---|---|
| Energy (kcal) | 2,000 | 1 | 2 | 2 | Equality (fixed) |
| Protein (g) | 50 | 0.025 g/kcal | 2 | 1.5 | Adequacy |
| Carbohydrate (g) | 275 | 0.138 g/kcal | 1.5 | 1.5 | Adequacy |
| Total Fat (g) | 78 | 0.039 g/kcal | 1.5 | 1.5 | Adequacy |
| Fiber (g) | 28 | 0.014 g/kcal | 2 | 1 | Adequacy |
| Sodium (mg) | 2,300 | 1.15 mg/kcal | 1 | 3 | Upper bound |
| Saturated Fat (g) | 20 (10 % kcal) | 0.010 g/kcal | 1 | 3 | Upper bound |
| Added Sugars (g) | 50 (10 % kcal) | 0.025 g/kcal | 1 | 3 | Upper bound |
| Potassium (mg) | 4,700 | 2.35 mg/kcal | 2 | 1 | Adequacy |
| Calcium (mg) | 1,300 | 0.65 mg/kcal | 2 | 1 | Adequacy |
| Iron (mg) | 18 | 0.009 mg/kcal | 2 | 1 | Adequacy |
| Vitamin D (µg / IU) | 20 µg (800 IU) | 0.010 µg/kcal | 1.5 | 1 | Adequacy |
| Others (if present) | | | 1 | 1 | Neutral |

**Supplementary Table 10 | Nutrient weights and per-kcal target values used in RDI-per-kcal optimization.** Nutrient-specific weights ($w_k$, $v_k$) and normalized RDI-per-kcal targets used in the generator's objective function. Weights emphasize adequacy for key nutrients (energy, protein, fiber, potassium) and moderation for over-consumed ones (sodium, saturated fat, added sugars). RDI values correspond to FDA/USDA Daily Values for a 2,000 kcal reference diet. Per-kcal targets are scaled by meal-energy fractions before optimization. Together, these parameters guide the convex nutrient-anchored portioning that yields balanced, guideline-consistent meals.